\documentclass{article}
\usepackage{spconf,amsmath,graphicx}
\usepackage{url}
\usepackage{hyperref}
\usepackage{amssymb}
\usepackage{booktabs}
\usepackage{enumitem}
\setlist{nosep, leftmargin=14pt}

\usepackage{mwe} 

\title{Temporally Consistent and Controllable Video Generation of 2D Cine CMR via Latent Space Motion Modeling}
\name{Yiheng Cao$^1$*, Gustavo Andrade-Miranda$^2$, Jiatian Zhang$^1$, Guillaume Sallé, Xin Gao$^1$\thanks{*Corresponding author. \\ Animations and demo will be available on \url{https://github.com/cyiheng/TextToCine2DMRI}}}
\address{
$^1$ Suzhou Institute of Biomedical Engineering and Technology, Chinese Academy of Sciences, Suzhou, China \\ $^2$ SyCoIA, IMT Mines Ales, Ales, France
}
\begin{document}
%
\maketitle
\begin{abstract}

Cine cardiac magnetic resonance is the gold standard for assessing cardiac function, but the scarcity of public datasets limits the development of advanced data-driven models. To address this limitation, we propose a generative method for synthesizing temporally coherent and anatomically consistent cardiac sequences. Our text-to-video framework decouples cardiac spatial structure from temporal motion. First, a fine-tuned diffusion model synthesizes an initial frame from a clinical text prompt, controlling anatomical features. Then, a latent flow model conditioned on a cardiac phase embedding generates the complete cardiac motion, ensuring spatial consistency and temporal control. Our model generates anatomically and pathologically diverse sequences with high temporal coherence and strong fidelity to input prompts, achieving a FID of 31.68 for image realism and a CLIP score of 31.04 for text-image alignment. These experimental results highlight its potential to produce high-fidelity, on-demand medical data, offering a scalable solution to data scarcity.

\end{abstract}
\section{Introduction}
\label{sec:intro}

Cardiovascular disease remains a leading cause of mortality worldwide, underscoring the critical importance of early and accurate diagnosis. \cite{global_burden_of_cardiovascular_diseases_and_risks_2023_collaborators_global_nodate}. Cardiac magnetic resonance (CMR) imaging, particularly cine sequences, has emerged as a gold-standard non-invasive modality for assessing cardiac function and morphology \cite{pennell_clinical_2004}. These dynamic acquisitions allow clinicians to visualize the structure and motion of the heart throughout the cardiac cycle. The detailed anatomical and functional information provided by cine CMR is essential to diagnose a wide range of cardiovascular pathologies \cite{karamitsos_role_2009,maron_hypertrophic_2013}.

\begin{figure}[htb]

  \centering
  \centerline{\includegraphics[width=\columnwidth]{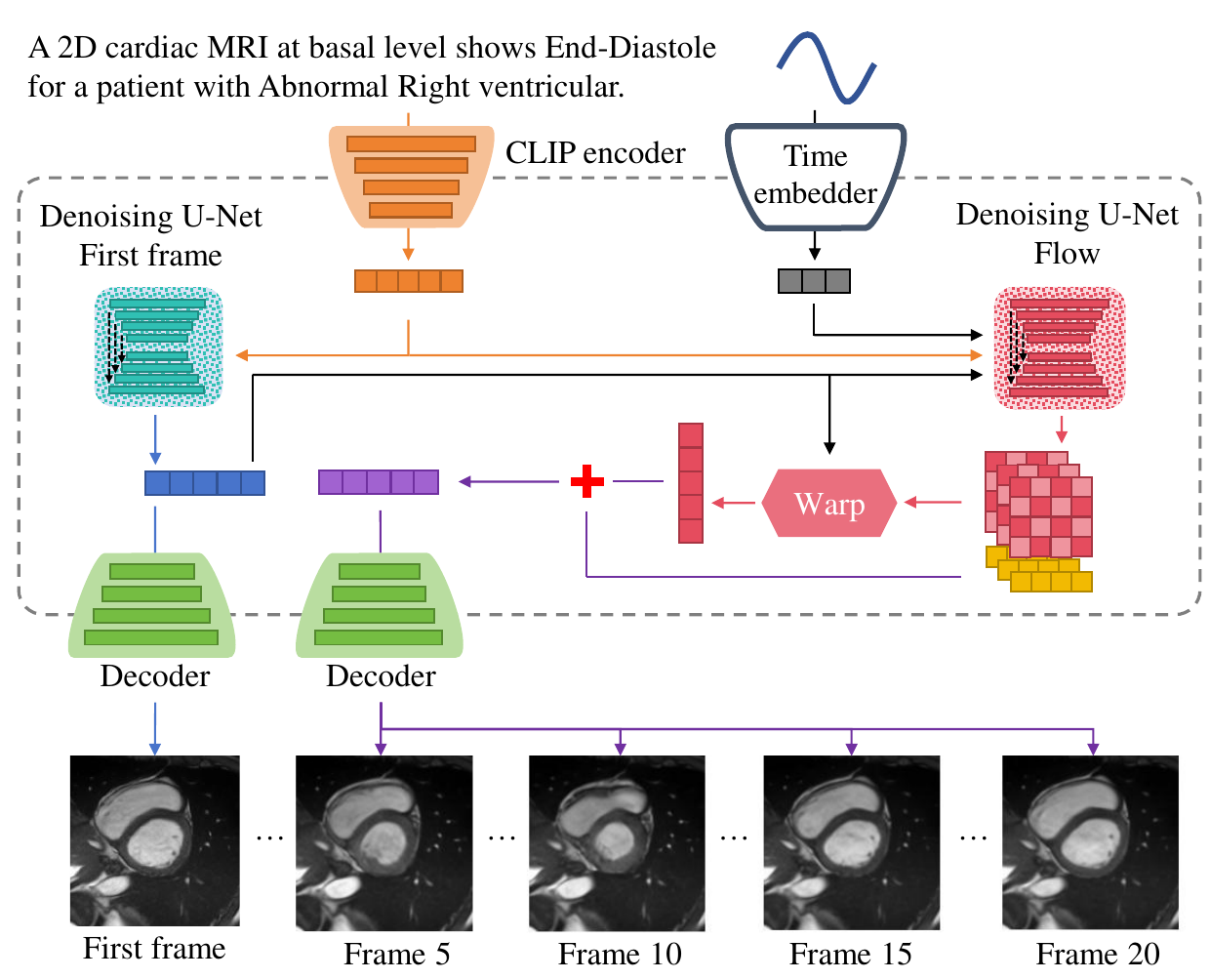}}
\caption{Overview of the motion sequence generation pipeline. The text prompt is used to generate the first frame, which—together with its latent vector and a sinusoidal time signal—conditions the flow generator. The resulting flows warp the initial latent representation, and the decoder reconstructs the motion sequence frames.}
\label{fig:overview}
\end{figure}

The quantitative analysis of cine CMR is traditionally time-consuming and subject to inter-observer variability. However, advances in deep learning (DL) have enabled automation and standardization of this process while maintaining high precision \cite{bai_automated_2018}. Deep learning-based approaches have demonstrated remarkable success in tasks such as cardiac segmentation and disease classification, often achieving performance comparable to that of human experts, while significantly improving speed and reproducibility \cite{petersen_reference_2017,chen_deep_2020}.

Despite these advantages, the performance and generalization abilities of DL models remain highly dependent on the availability of large, diverse, and annotated datasets. However, publicly available datasets are often insufficient, being limited in both size and pathological diversity \cite{litjens_survey_2017}. This lack of heterogeneity across patient populations, imaging hardware, and acquisition protocols is a major source of domain shift that compromises model robustness \cite{guo_impact_2024}. To mitigate data scarcity, data augmentation techniques are commonly employed. However, traditional augmentation methods, which are typically based on simple geometric and global intensity transformations, fail to produce novel or pathologically realistic examples, providing limited benefit in mitigating intrinsic dataset biases \cite{shorten_survey_2019}. Recently, diffusion-based models have shown promise to generate plausible synthetic images for data augmentation. Nevertheless, most existing applications remain constrained to static image domains or lack explicit temporal control \cite{Cheng_ISBI_2024,bluethgen_visionlanguage_2025}.

This study introduces a text-to-video framework to generate realistic 2D cine CMR sequences to address data scarcity. The proposed method fine-tunes a stable diffusion (SD) model to synthesize high-quality end-diastolic (ED) frames from clinical text prompts. Subsequently, a latent flow model generates the complete dynamic sequence by warping the latent representation of the initial frame, thereby simulating the complete cardiac motion. This process is guided by a controllable temporal signal to ensure temporal coherence. Moreover, by warping the initial frame's latent representation, the flow model preserves the underlying spatial structure, avoiding the stochastic variations that can arise from generating each frame independently with the diffusion model. Experiments on public datasets demonstrate that the proposed model produces a diverse range of anatomically and pathologically plausible sequences, establishing it as a valuable tool for synthetic data generation. 

\section{Methods}
\label{sec:methods}

Our proposed text-to-video framework generates 2D cine CMR sequences in a multi-stage process. The complete pipeline consists of four stages: (1) fine-tuning a variational autoencoder (VAE) to the CMR domain, (2) training a conditional diffusion model to synthesize a first frame from text prompt, (3) training a latent flow model to warp the reference frame to a target frame in the latent space, and (4) training a second diffusion model to generate the latent motion field. These components are trained sequentially and then integrated into a unified inference pipeline. 

\begin{figure}[ht]
    \centering
    \includegraphics[width=\columnwidth]{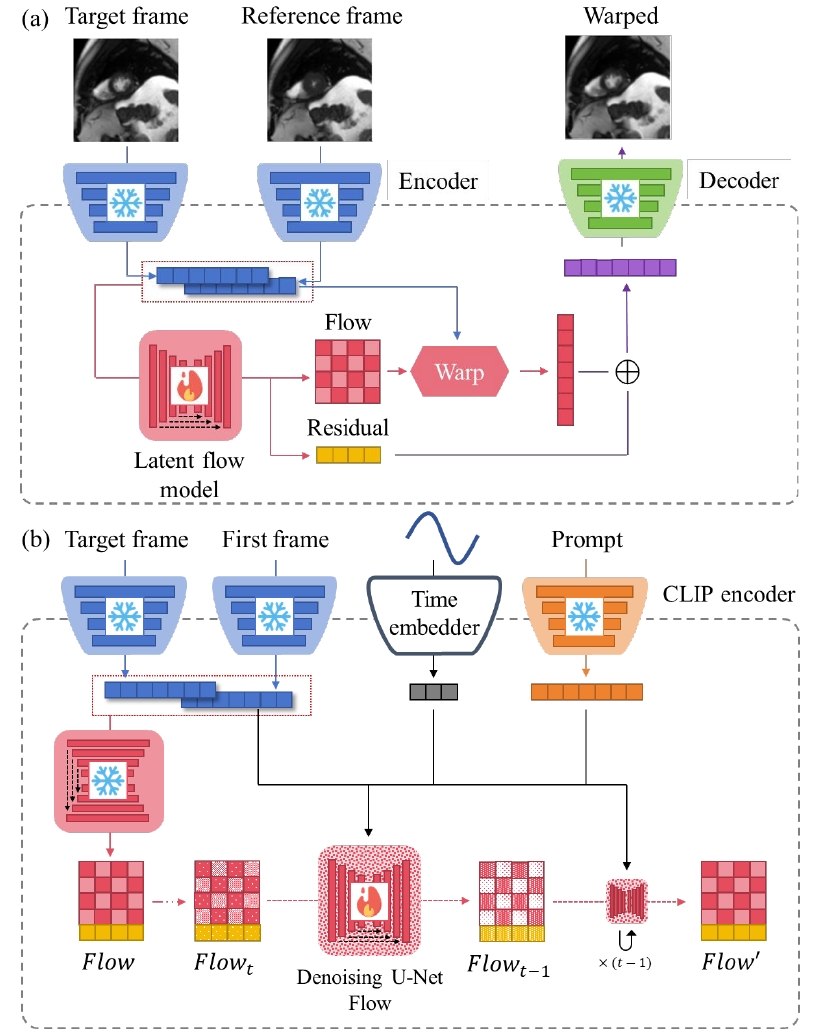}
    \caption{The two training stages of our latent flow model. (a) First, a deterministic network is trained to calculate the motion flow between a reference and target latent. (b) Then, a diffusion model learns to generate this flow, using the first frame, a text prompt, and a time signal for guidance.}
    \label{fig:pipeline}
\end{figure}

\subsection{Initial frame generation}
We adapt the latent space of a SD model to the grayscale 2D CMR domain by fine-tuning its VAE component. Then, to generate the initial ED frame of the cine sequence, we fine-tune the denoising U-Net backbone of the SD model. This model is trained to denoise a latent vector $z_t$ at timestep $t$ based on a conditioning input $c$. 
The conditioning $c$ is a text prompt containing clinical information such as anatomical level (e.g., "basal", "apical"), diagnosis (e.g., "hypertrophic cardiomyopathy"), and patient demographic information. This enables the generation of an anatomically plausible and context-specific first frame from a descriptive text input during inference.

\subsection{Deterministic latent flow}

To create a temporally coherent video sequence, we model the cardiac motion as a transformation within the latent space. We first train a U-Net to estimate the motion between a reference latent $z_{ref}$ and a target latent $z_{tgt}$ from different time point. The network takes the latents $(z_{ref},z_{tgt})$ as inputs and predicts two components: a 2D flow field $\phi$ and a latent residual map $r$. The flow field $\phi$ warps the reference latent to align it with the target, $z_{pred} = warp(z_{ref}, \phi)$, thereby capturing the primary motion. The residual map $r$ compensates for disocclusions and subtle appearance variations not captured by the 2D warping operation alone (Fig.~\ref{fig:pipeline}a).


\subsection{Latent flow diffusion model}

The deterministic model requires a target latent $z_{tgt}$ to estimate motion, which is not available during inference. To address this limitation, we introduce a second diffusion model trained specifically to generate both the 2D latent flow field $\phi$ and the residual map $r$. This motion model is conditioned by three inputs: (i) the latent representation of the initial frame, $z_0$; (ii) the text prompt; (iii) a temporal phase embedding, a normalized signal indicating the position of the target within the cardiac cycle (Fig.~\ref{fig:pipeline}b).

During the final inference stage, we first generate the latent representation of the initial frame $z_{0}$. For each subsequent timestep in the sequence, the motion diffusion model receives $z_{0}$, the text prompt, and the corresponding temporal phase embedding as inputs. The model then generates the latent motion field $\phi_t$, which is used to warp $z_{0}$ into the latent representation of the target frame, $z_{t}$. Finally, the VAE decoder reconstructs the corresponding pixel-space image, resulting in a complete, dynamically consistent cardiac cine sequence (Fig.~\ref{fig:overview}).

\section{Experiments and Results}
\label{sec:pagestyle}

\begin{table}[t]
  \caption{Quantitative evaluation of the fine-tuned VAE and the latent flow model}
  \centering
  \resizebox{0.95\linewidth}{!}{
  \begin{tabular}{@{}cc|cc@{}}
  \toprule
  \toprule
     & VAE              & \multicolumn{2}{c}{Latent flow model} \\ \midrule
     & Reconstruction   & ED to ES          & ES to ED          \\ \midrule
MSE $\downarrow$ & 27.11 $\pm$ 3.60 & 30.86 $\pm$ 17.05 & 28.96 $\pm$ 14.73 \\
PSNR $\uparrow$ & 34.74 $\pm$ 4.24 & 34.40 $\pm$ 4.85  & 34.63 $\pm$ 4.77  \\
SSIM $\uparrow$ & 0.93 $\pm$ 0.02  & 0.92 $\pm$ 0.02   & 0.93 $\pm$ 0.02   \\ \bottomrule\bottomrule
  \end{tabular}
  }
  \label{table:recons}
\end{table}

\begin{figure}[b]
    \centering
    \includegraphics[width=0.9\columnwidth]{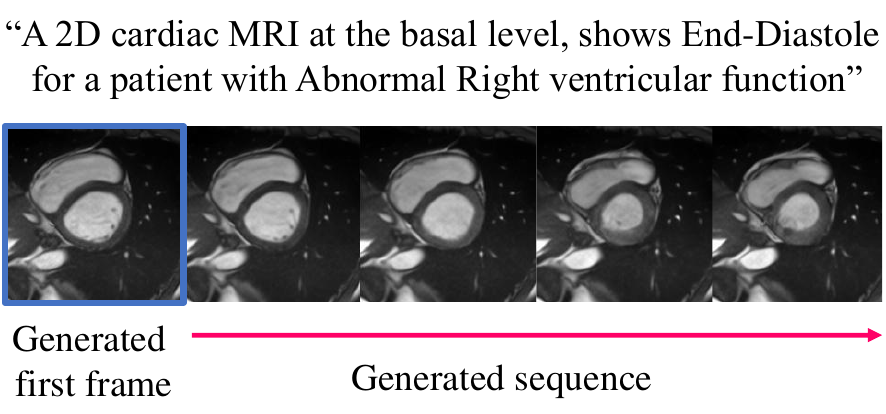}
    \caption{Qualitative results of our text-to-video generation pipeline. The model synthesizes the initial frame (blue box) based on the prompt, then generates the subsequent motion. }
    \label{fig:seq}
\end{figure}

\subsection{Experimental setup}

We trained and evaluated our model on two public datasets: the Automated Cardiac Diagnosis Challenge (ACDC) \cite{bernard_deep_2018} and the Kaggle Second Annual Data Science Bowl (DSB) \cite{kaggle}. All images were sliced into 2D frames and center-cropped to a resolution of 192$\times$192 pixels. For latent flow training, frame pairs were selected with a temporal distance ranging from 2 to 12 frames. Text prompts were constructed from patient metadata and anatomical information. Temporal conditioning was achieved using a phase embedding derived from the relative timing between ED and end-systolic (ES) frames. All models were trained on a single NVIDIA RTX 4090 GPU.

To evaluate performance, we assessed the VAE and latent flow model using the Mean Squared Error (MSE), Structural Similarity Index (SSIM), and the Peak Signal-to-Noise Ratio (PSNR). For the text-conditioned first frame, we used the Fréchet Inception Distance (FID) and CLIP score to measure image fidelity and text-alignment, respectively. To evaluate temporal coherence, we computed the Content-Debiased Fréchet Video Distance (FVD) \cite{ge2024content} using an I3D backbone. CD-FVD was selected to disentangle temporal dynamics from spatial appearance, strictly evaluating the generated cardiac motion. We compared our model against a 'Real VS Real' oracle baseline to establish a lower bound.

\subsection{Quantitative and qualitative evaluation}



The Table~\ref{table:recons} shows that the latent flow model's ability to warp frames between ED and ES phases achieves SSIM scores ($0.92$ and $0.93$) that are similar to the VAE's direct reconstruction ($0.93$). The MSE and PSNR also show similar trend. This indicates that our approach preserves structural integrity and image quality with small degradation while capturing the motion of the cardiac cycle.

\begin{table}[]
  \centering
  \caption{Quantitative evaluation of generation quality.}
  \resizebox{0.95\linewidth}{!}{
  \begin{tabular}{@{}ll|ll@{}}
  \toprule
  \toprule
            & First frame &                             &  Motion \\ \midrule
  FID $\downarrow$        & 31.68                 & FVD $\downarrow$ (Real VS real)           & 19.41             \\
  CLIP score $\uparrow$  & 31.04 $\pm$ 1.38      & FVD $\downarrow$ (Real VS fake)      & 317.94            \\ \bottomrule\bottomrule
  \end{tabular}
  }
  \label{table:ff}
\end{table}

As shown in Table~\ref{table:ff}, our text-conditioned first-frame generator achieves an FID of $31.68$, indicating that the synthesized images are both realistic and anatomically plausible. A high CLIP score of $31.04$ further confirms accurate semantic alignment with the text prompt, demonstrating precise control over the generated content.

To assess motion, we computed FVD on sequences covering the full cardiac cycle. We report a CD-FVD of $317.94$ against a real-data baseline of $19.41$. While a gap remains compared to real data, this metric confirms that the model captures complex, non-linear cardiac dynamics consistent with the learned distribution, rather than producing static or incoherent sequences.

Fig.~\ref{fig:seq} displays a 5-frame sequence generated by the model. 
The sequence demonstrates high visual fidelity and temporal coherence, depicting a smooth and realistic cardiac cycle of contraction and relaxation. The anatomy is consistent with the specified basal view, and the model accurately interprets the pathological prompt to generate the abnormal right ventricular function. These results highlight the framework's ability to produce anatomically correct and clinically meaningful dynamic sequences from text descriptions.

\section{Conclusion and Discussion}
\label{sec:typestyle}

In this work, we introduced a framework for generating controllable 2D cine CMR sequences from text. Our approach successfully models cardiac motion, producing temporally coherent sequences with high fidelity. A key strength is the use of a latent flow model guided by a controllable temporal signal. This approach is crucial for maintaining anatomical integrity, avoiding irregular structural variations that can arise from independent, per-frame generation. Our initial focus on 2D data served as a proof-of-concept, validating our methodology while remaining within manageable computational resource limits.  Although our multi-stage pipeline introduces a risk of error propagation, it decouples the computational load, enabling training and inference on consumer-grade GPUs where unified video models often exceed memory constraints. However, this 2D approach does not explicitly account for through-plane motion or spatial continuity across the z-axis, and the simplified temporal signal does not fully capture complex physiological rhythms. Future work will focus on extending the framework to 3D to generate complete cardiac volumes and use realistic signal conditioning, such as electrocardiogram. Finally, to establish the clinical value of our approach, we will quantify the benefit of using our generated sequences for data augmentation in downstream tasks such as cardiac segmentation and disease classification.



\section{Compliance with Ethical Standards}
\label{sec:ethical}

This research study was conducted using medical imaging data made available in open access by \href{https://www.creatis.insa-lyon.fr/Challenge/acdc/databases.html}{CREATIS}\cite{bernard_deep_2018} and \href{https://www.kaggle.com/competitions/second-annual-data-science-bowl/overview}{Kaggle}\cite{kaggle}. Ethical approval was not required as confirmed by the license attached with the open access data.

\section{Acknowledgments}
\label{sec:acknowledgments}

This work is supported by the Research Fund for International Young Scientists of the National Natural Science Foundation of China (No. W2533209)

\bibliographystyle{IEEEbib}
\bibliography{refs}

\end{document}